\documentclass[conference]{IEEEtran}
\IEEEoverridecommandlockouts
\usepackage{cite}
\usepackage{amsmath,amssymb,amsfonts}
\usepackage{hyperref}
\usepackage{algorithmic}
\usepackage{graphicx}
\usepackage{subfig}
\usepackage{textcomp}
\usepackage{xcolor}
\def\BibTeX{{\rm B\kern-.05em{\sc i\kern-.025em b}\kern-.08em
    T\kern-.1667em\lower.7ex\hbox{E}\kern-.125emX}}
\begin{document}

\title{Detecting Abnormal Health Conditions in Smart Home Using a Drone}

\author{\IEEEauthorblockN{Pronob Kumar Barman}
\IEEEauthorblockA{\textit{Department of Information Systems} \\
\textit{University of Maryland Baltimore County}\\
Baltimore, USA }

}
\maketitle

\begin{abstract}
Nowadays, detecting aberrant health issues is a difficult process. Falling, especially among the elderly, is a severe concern worldwide. Falls can result in deadly consequences, including unconsciousness, internal bleeding, and often times, death. A practical and optimal, smart approach of detecting falling is currently a concern. The use of vision-based fall monitoring is becoming more common among scientists as it enables senior citizens and those with other health conditions to live independently. For tracking, surveillance, and rescue, unmanned aerial vehicles use video or image segmentation and object detection methods. The Tello drone is equipped with a camera and with this device we determined normal and abnormal behaviors among our participants. The autonomous falling objects are classified using a convolutional neural network (CNN) classifier. The results demonstrate that the systems can identify falling objects with a precision of 0.9948.
\end{abstract}

\begin{IEEEkeywords}
Ryze Tello, CNN, Fall detection
\end{IEEEkeywords}

\section{Introduction}
 According to the World Health Organization (WHO), over 680,000 fatal falls occur worldwide each year, with the majority of victims being individuals over the age of 65\cite{a1}. Falls are the leading cause of fatal injuries in older adults and are especially dangerous for those living independently. In order to address this issue, the purpose of our study is to detect falls within the home. With this, we hope to reduce the number of deaths associated with falls and allowing the elderly and others that are prone to falls to still live independently. 
 \\\\
 An unmanned aerial vehice (UVA), often referred to as a drone is an aircraft without any humans on board. Drones are often used in a number of situations such as emergency rescue, asset protection, and wildlife monitoring, but they can do more. Introducing a drone into the home increases vigilance that allows victims to receive the help that they need in an appropriate amount of time. More specifically, the Tello drone consists of a built-in camera that will be used for tracking, surveillance, and rescue. Through our study, we believe that a drone can be an effective smart home device to help reduce mortality rates and the need of in-home aids for the elderly and others prone to falls. 
\\
 \subsection{Proposed Approach}\label{AA}
Although our current solution only solves the fall detection aspect, we envisioned our project to have an end-to-end solution based on person detection and fall classification. 
\\\\
As seen in Figure \ref{fig1} below, our solution accounts for all possible outcomes of person detection and fall classification. The drone will constantly be scanning the area and taking images and videos and if a person is detected we move on to feature extraction to gather all relevant information on the human's movement. If a fall is detected the user will be asked whether or not they need help. If the human answers yes or does not give an answer at all, an alarm will be triggered, contacting emergency personnel along with family members programmed into the system. However, if a person or a fall is not detected, the drone will go back to the beginning phase to continue its analysis. Lastly, if there is any uncertainty, the drone will re-evaluate by moving to a new location to reassess the situation. 
\\\\
As previously mentioned, our current solution does not offer an end-to-end solution, but this serves as the framework we envisioned our project to follow.

\begin{figure}[htbp]
\centerline{\includegraphics[scale=0.45]{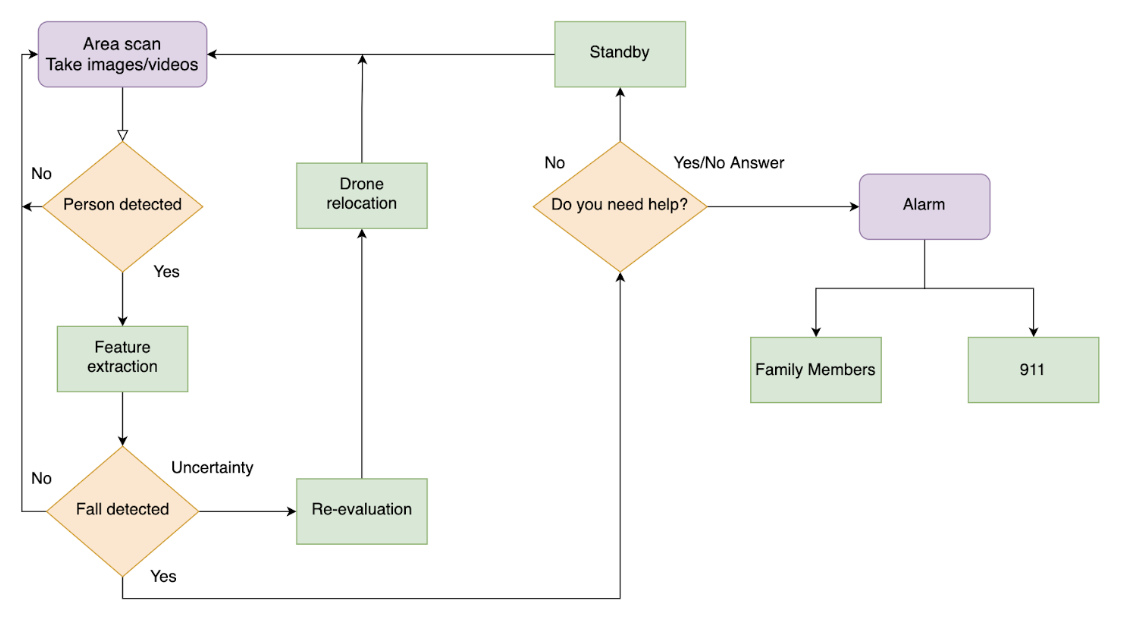}}
\caption{Proposed Approach Flowchart}
\label{fig1}
\end{figure}

\subsection{Related Work}
Previous studies have presented fall detection systems used to detect patient falls. While other studies have designed a UAV to monitor patients' behavior in order to alert emergency personnel for victims of falls. This section covers previous work related to UAVs and other fall detection systems such as robots.
\\
\subsubsection{Work Related to UAV Systems}
Previous studies have designed UAVs to transport medical supplies and medication to patients\cite{a2}. Other studies have adopted UAVs to have an advanced first aid system to monitor elderly patients with heart conditions putting them at risk of falls. These authors proposed a wearable sensor-based fall detection device (FDD) designed for monitoring HR and detecting falls, and a call emergency center (CEC) to receive messages from the FDD and plan the UAV's path\cite{a3}. The FDD is attached to the patient's upper arm and once it detects a fall and the HR is measured as abnormal, the FDD sends a message to the CEC which includes the patient's personal information. The advanced first aid system provides first aid to the patients that are prone to falling\cite{a3}. The first aid package is then delivered to the patient via the UAV based on the GPS coordinates of the FDD. 
\\\\
The results illustrated excellent performance where the proposed FDD had an accuracy of 99.2\% when distinguishing between falls and normal activities. Furthermore, in terms of HR monitoring, results show that the proposed algorithm for the FDD was successful and can be used for monitoring the elderly who have atrial fibrillation (irregular heartbeat), which causes falls\cite{a3}. Moreover, the GPS was accurate at determining the patient's location at the time of the fall. 
\\\\
In the future, researchers are interested in the development of the entire system, including the design of a system that works autonomously to receive patient information and send a UAV without human intervention\cite{a3}.
\\
\subsubsection{Work Related to Robots}
Previous studies have developed mobile robots that monitors one person in a single view with constant supervision, while other studies proposed a more robust system. Saturnino Maldonado-Bascon et al., proposed autonomous mobile-patrol robots that integrate privacy protection and only necessary supervision. The robot was designed to autonomously patrol an indoor environment and when a fall is detected, an alarm is activated\cite{a4}. This study also believes that vision-based systems offer more advantages than wearable sensor-based systems. However, in vision-based systems, cameras play a critical role. In order to solve the problem of occlusion, the system should be equipped with multiple cameras in order to monitor and assess a number of angles\cite{a4}. Recent advancements in Computer Vision support this idea more boldly\cite{mkcnn,mk,maloy}. This study combines the YOLOv3 algorithm based on a convolutional neural network (CNN) for person detection and a support vector machine (SVM) for fall classification\cite{a4}. The basis for this study's fall detection approach was an assistance robot named LOLA, that was designed entirely by the research team. This device was developed to monitor and help the elderly and others with functional disabilities living alone. Not only is LOLA an assistive robot, but it also serves as a rollator for assistance with walking or as a table\cite{a4}. 
\\\\
When evaluating the performance of the end-to-end solution, results show that the fall classifier detected 390 true positives, 1 false negative, and 0 false positives, indicating values of 100\% and 99.74\% for precision and recall, respectively\cite{a4}. Overall, results indicate that the system had a high success rate in fall detection. 
\\\\
In the future, researchers hope to investigate and improve upon occlusion detection and possibly merging person detection and fall classification into one CNN\cite{a4}. 
\\
 \section{Data Collection}
 In our study we utilized two datasets. The first dataset is available to the public and was used to train our model. The second dataset is our primary dataset that was collected from two male participants aged 12 and 16 years old. We collected the data in two locations:
 \begin{itemize}
     \item living room one
     \item living room two
 \end{itemize}
Furthermore, we followed IRB standards since we were collecting data from human participants. 
 
\begin{figure}[ht]
  \centering
  \subfloat[Not Fall]{\includegraphics[width=0.2\textwidth]{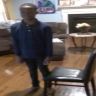}\label{fig:f1}}
  \hfill
  \subfloat[Fall]{\includegraphics[width=0.2\textwidth]{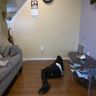}\label{fig:f2}}
  \caption{Sample Images of Fall Detection}
\end{figure}

\subsection{Design of Hardware}\label{AA}
We used a Ryze Tello drone as a low-cost hardware platform to perform our study. The Ryze Tello drone costs approximately 100 USD and is a small and light weight drone that weighs approximately 80 g (with propellers and battery). The full specifications are given below:\\
\begin{itemize}
    \item Dimensions: 98×92.5×41 mm
    \item Propeller: 3 inches
    \item Built-in Functions: Range Finder, Barometer, LED, Vision System, 2.4 GHz 802.11n Wi-Fi, 720p Live View
    \item Port: Micro USB Charging Port
    \item Detachable Battery: 1.1Ah/3.8V
\end{itemize}

\begin{figure}[htbp]
\centerline{\includegraphics[scale=0.45]{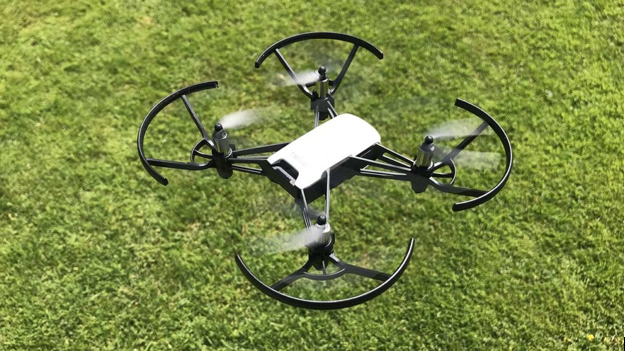}}
\caption{Ryze Tello Drone}
\label{fig3}
\end{figure}

The drone can be operated both indoors and outdoors. A single camera, an IMU with a 3-axis gyroscope, 3-axis accelerometer, and 3-axis magnetometer, and a pressure and IR-based altitude sensors are all incorporated into the drone. The front-facing camera features an 8-megapixel sensor and can record video at a resolution of $1280 \times 720$ at 30 frames per second.
\\
\subsection{Device Setup and Requirements}\label{AA}
The Tello drone is a ready-to-use device that requires no additional setup. In order to control the drone, a compatible mobile phone is required along with the 'Tello' app and wifi. The drone is a battery charged device, lasting approximately 12-13 minutes. 
\\
\subsection{Study Design}\label{AA}
As previously mentioned, our study consisted of two male participants, aged 12 and 16 years old. This study took place in a single-family home with approximately 10 inches ceilings. The images and videos were captured in two different rooms within the home, one with more open space and one with less space. The study was carried out over a span of two weeks, with sessions ranging from 15 minutes to an hour depending on drone complications, participant poses and assessments, and retakes. 
\\
\subsection{Participant Instructions}\label{AA}
The instructions that our participants followed were straightforward. Using our own intuition, we first determined what positions people usually are in after a fall and took note of it. We then showed our participants an image of the position and instructed them to reenact it. Once our participants were in position, we powered up the drone, captured the videos and images, assessed them and then asked our participants to move on to the next position. 
\newpage
\section{Proposed Methodology}
In this section we briefly discuss fall detection systems and methods. 
\\
\subsection{Convolutional Neural Network (CNN)}
We propose a deep learning-based approach for detecting whether or not a person has fallen using RGB photographs. For feature extraction, we employ a CNN algorithm\footnote{The proposed idea is taken from \cite{a8}} with 6 blocks and a header. Conv2D Layer with ReLU activation and MaxPooling2D are included in each block. We begin with 16 troops and increase by 2 units every 2 blocks:
$$16 -> 16 -> 32 -> 32 -> 64 -> 64$$
\\
We flatten the result after feature extraction, add a Dense layer of 32 units, and utilize sigmoid activation for classification.
\\\\
The CNN model uses visual representations based on
the RGB image and segmentation and learns high-level embedding features for fall recognition. We explain the individual components of the conceptual structure in detail below. 
\begin{figure}[htbp]
\centerline{\includegraphics[scale=0.3]{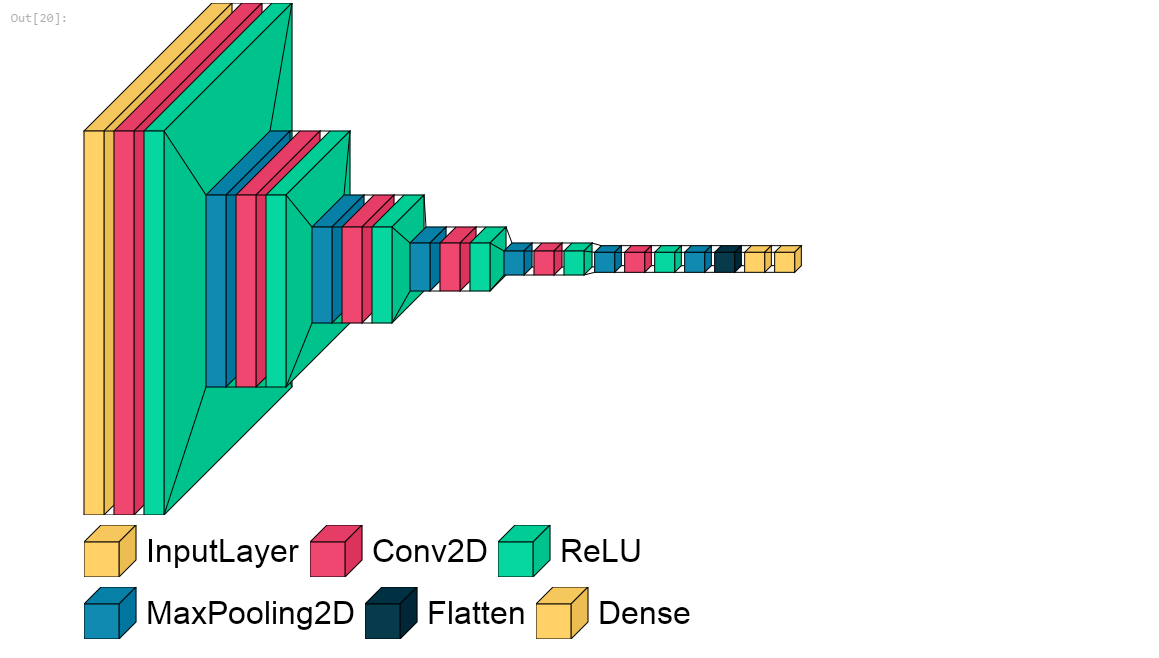}}
\caption{CNN Architecture \cite{a8}}
\label{fig4}
\end{figure}

\begin{figure}[htbp]
\centerline{\includegraphics[scale=0.5]{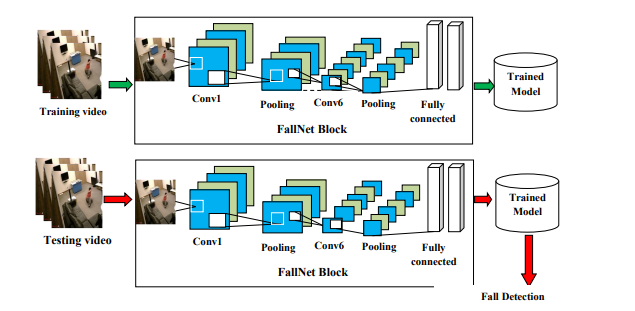}}
\caption{Overall Architecture\cite{a5}}
\label{fig5}
\end{figure}
\newpage
\subsection{Training and Implementation}
\begin{figure}[htbp]
\centerline{\includegraphics[scale=0.7]{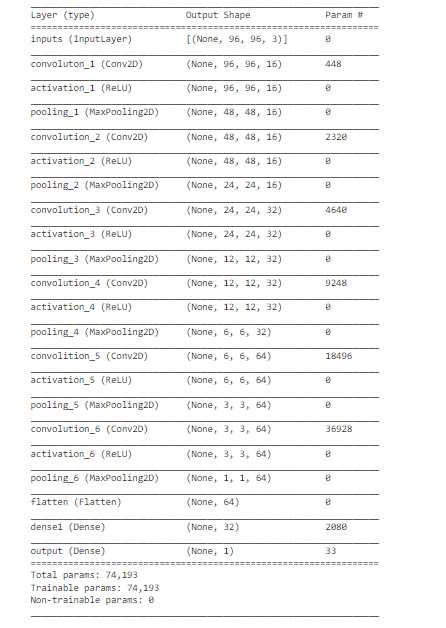}}
\caption{Summary of the FallNet Model \cite{a8}}
\label{fig6}
\end{figure}

Initialize the weights of the nodes in the FallNet network model at the start. The weights for the convolutional layer are started in this model using the weights from the following embedding layer, which have a zero-mean gaussian distribution (SD= 0.01 and bias=0) \cite{a5}. We utilize the Adam optimizer with a learning rate of 0.0001 and a loss function of Binary Crossentropy to train our model. The proposed system is built on the basis of Torch, Keras libraries\cite{a6,a7}. Figure \ref{fig6} depicts the convolution layers, which range from 1-6 convolutions.

\section{Experimental Results}
\subsection{Datasets}
At first, we utilized this publicly available dataset \href{http://fenix.univ.rzeszow.pl/~mkepski/ds/uf.html}{data} to train our model for fall detection. The acquired data was then used to validate the pre-trained model. The accuracy and loss plot was obtained after 5 epochs.
\\
\subsection{Evaluation Metrics}
The loss is the value that a neural network tries to minimize in deep learning: it's the difference between the ground truth and the forecasts. The neural network learns to decrease this distance by altering weights and biases in a way that minimizes the loss.
\\\\
In regression problems, for example, you have a continuous target, such as height. The discrepancy between your predictions and the actual height is what you wish to minimize. You can use mean absolute error as a loss to tell the neural network what it should be minimizing.
\\\\
It's a little more complicated in terms of classification, but it's still comparable. Probability is used to predict classes and the neural network minimizes the possibility of assigning a low probability to the real class in classification. Typically, the loss is categorical crossentropy.
\\\\
The difference between loss and validation loss is that the former is applied to the train set, while the latter is applied to the test set. As a result, the latter is a useful indicator of how the model performs on data that hasn't been seen before. Use validation data=[$x_{test}, y_{test}$] or validation split=0.2 to get a validation set.
\\\\
To avoid overfitting, it's advisable to rely on the validation loss. When the model fits the training data too closely, overfitting occurs, and the loss continues to decrease while the val loss remains stale or grows. When the validation loss stops lowering, you can use EarlyStopping in Keras to halt training.
\\\\
After performing 5 epochs we get the following accuracy and loss plot from training and testing data.
\begin{figure}[htbp]
\centerline{\includegraphics[scale=0.7]{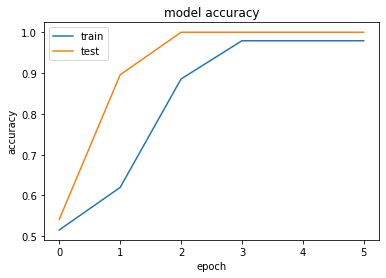}}
\caption{Accuracy of CNN Model}
\label{fig7}
\end{figure}

\begin{figure}[htbp]
\centerline{\includegraphics[scale=0.7]{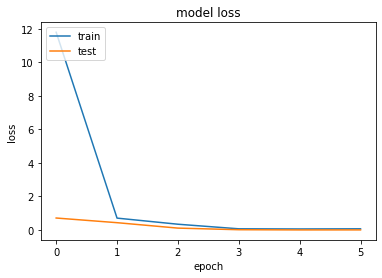}}
\caption{Loss Plot}
\label{fig8}
\end{figure}
\begin{figure}[htbp]
\centerline{\includegraphics[scale=0.27]{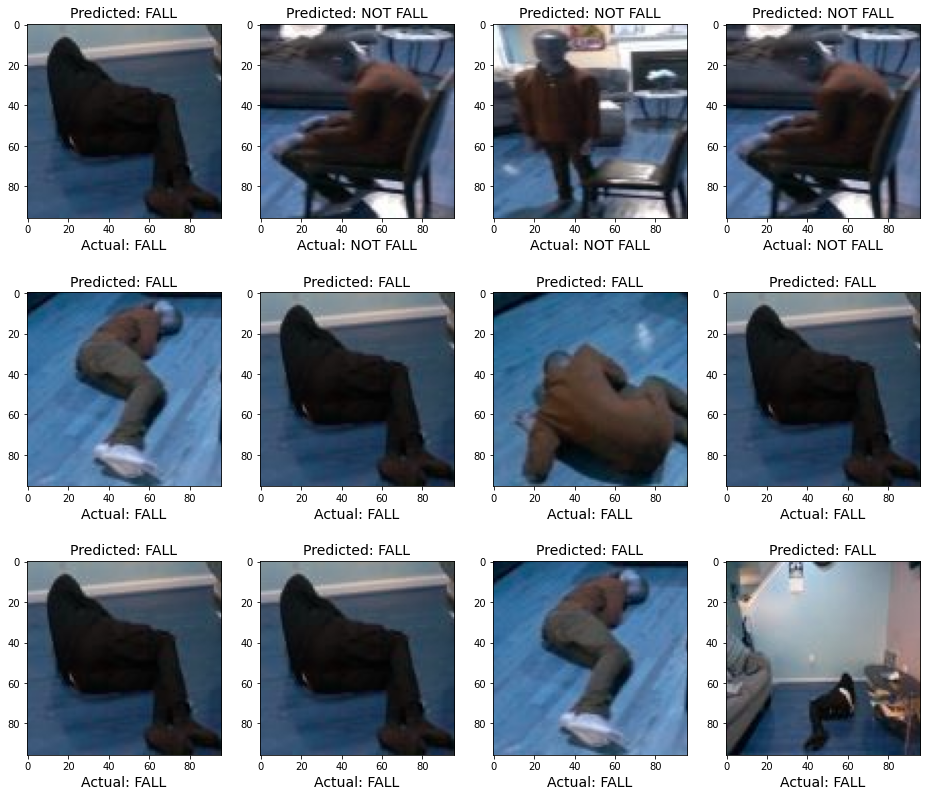}}
\caption{Model Prediction}
\label{fig9}
\end{figure}

\newpage
\begin{table}[htbp]
\centering
\caption{Summary Table}
\label{table:mylabel}
\begin{tabular}{|c|c|c|c|c|}
   \hline
    Epoch & Loss & Accuracy & Validation Loss & Validation Accuracy \\
    \hline
    $0$ & $18.1700$ & $0.4896$ & $0.7963$ & $0.4583$\\
    \hline
    $1$ & $0.6882$ & $0.5573$ & $0.5742$ & $0.8750$\\
    \hline
    $2$ & $0.4788$ & $0.8073$ & $0.1549$ & $0.9792$\\
    \hline
    $3$ & $0.1305$ & $0.9531$ & $0.0089$ & $1.0000$\\
    \hline
    $4$ & $0.0565$ & $0.9792$ & $0.0018$ & $1.0000$\\
    \hline
    $5$ & $0.0115$ & $0.9948$ & $0.000012$ & $1.0000$\\
    \hline
\end{tabular}
\end{table}
Table \ref{table:mylabel} shows that the model achieves complete accuracy after 5 epochs. Our model performs better on the unseen data, as seen in figures \ref{fig7} and \ref{fig8}. On testing data, our model is more accurate. Furthermore, the loss of model is minimal. Figure \ref{fig9} also displays the prediction rate of $100\%$. That means our model accurately detects the fall.

\section{Conclusion}
\subsection{Challenges and Limitations}
During the duration of our study, we faced some challenges and had some limitations along the way. This may be due to the inexpensive nature of the Tello drone, however during the initial stages of understanding how to fly the drone, the propellers would often fall off on it's own. After flying the drone several times, we encountered takeoff complications, where the drone could not takeoff from the ground. Even after re-calibrating the device multiple times, we could not successfully address the issue, causing us to utilize a secondary Tello drone. 
\\\\
Secondly, many of the training datasets available publicly required permission to use and given the time constraint of the project, we could not rely on waiting to be approved for access. 
\\\\
Also, for a number of poses we wanted our participants to re-enact, we could not achieve certain angles based on the height limitations of the drone in order to ensure safe flying. 
\\\\
Lastly, the drone's camera quality often gave us issues when testing the images with the CNN model, sometimes being incapable of determining what the image consisted of. 
\\
\subsection{Takeaway}
We used a small drone to detect falls in this experiment. We wanted to test if the Tello drone could be effective in monitoring smart health on a broader scale. Based on our findings, we can safely conclude that we can deploy a Tello drone to monitor the residence. There are some drawbacks as well, because the CNN model requires a large dataset with uniformly distributed classes. Also, the drone's range is limited, and the video resolution is only 720p with an 8-megapixel sensor. The weather circumstances contribute to hazy photos, and we are unable to fly it in strong winds. It is, nevertheless, quite inexpensive and convenient. We feel that this drone can still be used to monitor the residence in an interior scenario.
\\
\subsection{Learning Experience}
As machine learning and smart home devices, such as a drone are new topics that we had not previously explored, we learned a great deal throughout the duration of the project. First and foremost, it was critical to work together as a team from start to finish. We played on our strengths and assisted the other when in need of help. Furthermore, we learned the importance of trouble shooting because although technology is revolutionary, there will still be some challenges along the way. Being able to turn to one another and other resources when needed allowed us to achieve the best results that we could.
\\\\
In terms of the drone, we determined that a drone can be an effective device to detect falls in the home if created in a robust fashion. Although drones are initially thought of to be used in the entertainment industry to capture footage in areas where humans cannot reach, it can be applied to many other disciplines such as healthcare. A drone can be used in the home to detect falls, minimize the need for an in-home aid, and reduce unwanted stress.
\\\\
This project serves as an eye opener that smart home technology can be adapted to make the world a more safer place. 
\\
\subsection{Future Work}
In the future, we hope to explore whether the drone is capable alerting appropriate officials in the event of an incident. Furthermore, we would be interested in exploring the possibilities of having an autonomous drone or autonomous vehicle, taking away the burden of having to fly the drone manually. We aim to monitor the system for potential attacks and devise strategies to mitigate them, recognizing that autonomous drones or vehicles are susceptible to security threats\cite{maloy_ggnb,gcnids}. Furthermore, we would like to introduce the idea of having the drone dock itself when it needs to be charged. Lastly, since we did not include mask R-CNN in our analysis, comparing mask R-CNN and CNN algorithms in the future can prove beneficial to the effectiveness of our system.
\\
\section*{Acknowledgment}
I would like to express my deepest gratitude and appreciation to Professor Dr. Nirmalya Roy for his invaluable guidance, mentorship, and expertise throughout the course of this research endeavor. Furthermore, I extend my heartfelt thanks to Mariam Yekini for her diligent and meticulous data collection efforts. Mariam's dedication, attention to detail, and competence played a crucial role in ensuring the quality and accuracy of the data used in this study. Her tireless work was instrumental in the successful execution of this research project.
\\

\end{document}